\documentclass[10pt,journal]{IEEEtran}

\usepackage{amsmath,amssymb,amsthm}

\usepackage{array}
\usepackage{booktabs}
\usepackage{makecell}
\usepackage{multirow}
\usepackage[table]{xcolor}

\usepackage{graphicx}
\graphicspath{{pictures/}}
\usepackage[caption=false,font=footnotesize]{subfig}

\usepackage{algorithm}
\usepackage{algpseudocode}

\usepackage{cite}
\usepackage{url}

\usepackage{microtype}
\usepackage[hidelinks]{hyperref}

\theoremstyle{definition}

\newcommand{\AlgInput}[1]{\Statex \textbf{Input:} #1}
\newcommand{\AlgOutput}[1]{\Statex \textbf{Output:} #1}
\newcommand{\AlgPhase}[1]{\Statex \textit{#1}}

\makeatletter
\newcommand{\supp@origaddcontentsline}{}
\newcommand{\startsupplementarytoc}{%
  \let\supp@origaddcontentsline\addcontentsline
  \renewcommand{\addcontentsline}[3]{\supp@origaddcontentsline{stc}{##2}{##3}}%
}
\newcommand{\printsupplementarytoc}{%
  \section*{\contentsname}
  \@starttoc{stc}%
}
\makeatother

\newcommand{\startsupplementarydocument}{%
  \clearpage
  \setcounter{page}{1}%
  \renewcommand{\thepage}{S\arabic{page}}%
  \setcounter{section}{0}%
  \setcounter{subsection}{0}%
  \setcounter{subsubsection}{0}%
  \setcounter{table}{0}%
  \setcounter{figure}{0}%
  \setcounter{equation}{0}%
  \setcounter{algorithm}{0}%
  \setcounter{theorem}{0}%
  \setcounter{tocdepth}{2}%
  \setcounter{secnumdepth}{3}%
  \renewcommand{\thesection}{S-\uppercase\expandafter{\romannumeral\arabic{section}}}%
  \renewcommand{\thesubsection}{\thesection.\arabic{subsection}}%
  \renewcommand{\thesubsubsection}{\thesubsection.\arabic{subsubsection}}%
  \renewcommand{\thetable}{S-\uppercase\expandafter{\romannumeral\arabic{table}}}%
  \renewcommand{\thefigure}{S\arabic{figure}}%
  \renewcommand{\theequation}{S\arabic{equation}}%
  \renewcommand{\thealgorithm}{S\arabic{algorithm}}%
  \renewcommand{\theHsection}{supp.\arabic{section}}%
  \renewcommand{\theHsubsection}{supp.\arabic{section}.\arabic{subsection}}%
  \renewcommand{\theHsubsubsection}{supp.\arabic{section}.\arabic{subsection}.\arabic{subsubsection}}%
  \renewcommand{\theHtable}{supp.table.\arabic{table}}%
  \renewcommand{\theHfigure}{supp.figure.\arabic{figure}}%
  \renewcommand{\theHequation}{supp.equation.\arabic{equation}}%
  \renewcommand{\theHalgorithm}{supp.algorithm.\arabic{algorithm}}%
  \renewcommand{\contentsname}{Supplementary Contents}%
  \startsupplementarytoc
  \twocolumn[%
    \begin{center}
      {\LARGE Supplementary Document of ``Structure-Conditioned Actor--Critic Branches for Quality-Diversity Reinforcement Learning''\par}
      \vspace{0.75em}
    \end{center}
  ]
}

\begin{document}

\title{Structure-Conditioned Actor--Critic Branches for Quality-Diversity Reinforcement Learning}

\author{Lianrong~Zuo, Peilan~Xu, \IEEEmembership{Member,~IEEE}, Yong~Liu, and Wenjian~Luo, \IEEEmembership{Senior Member,~IEEE}%
  \thanks{Lianrong Zuo, Peilan Xu, and Yong Liu are with the School of Artificial Intelligence, Nanjing University of Information Science and Technology, Nanjing 210044, China (e-mail: 202383460103@nuist.edu.cn, xpl@nuist.edu.cn, and 202412491463@nuist.edu.cn). (\textit{Corresponding author: Peilan Xu.})}%
  \thanks{Wenjian Luo is with Guangdong Provincial Key Laboratory of Novel Security Intelligence Technologies, Institute of Cyberspace Security, School of Computer Science and Technology, Harbin Institute of Technology, Shenzhen 518055, Guangdong, China (e-mail: luowenjian@hit.edu.cn).}%
}

\maketitle

\begin{abstract}
Quality-diversity reinforcement learning (QD-RL) aims to construct policy repertoires that contain both high-performing and behaviorally diverse policies.
Existing QD-RL methods mainly diversify policy instances after rollout evaluation or use learned value information to improve policy quality and behavior targeting, while the learning branches that generate candidate policies remain less explored.
This paper proposes SV-QD-RL, a structure--value coupled framework that represents each candidate as a structure-conditioned actor--critic branch.
Each branch contains an actor, a structural mask, a branch-specific critic, a replay state, and evaluation attributes including behavior, return, sparsity, and value profile.
The structural mask defines the actor subspace in which the branch learns, while the branch-specific critic and replay state shape its value-learning trajectory.
A branch-aware QD archive then evaluates and retains branches according to behavioral quality, structural footprint, and value-profile information.
Experiments on MuJoCo continuous-control tasks show that SV-QD-RL constructs policy repertoires with strong archive quality and behaviorally useful diversity.
Ablation and diagnostic analyses further indicate that structural conditioning, critic differentiation, and memory-consistent refinement make complementary contributions to behavioral specialization.
Schedule-aware repertoire evaluation shows that the learned archive provides selectable policy alternatives under changing behavior-level requirements.
These results suggest that coupling actor structure with branch-specific value learning is an effective mechanism for generating diverse QD-RL policy repertoires.
\end{abstract}

\IEEEpeerreviewmaketitle

\section{Introduction}\label{sec:introduction}

\IEEEPARstart{R}{einforcement} learning (RL) commonly optimizes a single policy to maximize the expected return under a given task objective, which is effective when the desired behavior and operating condition are fixed.
However, a single learned policy may become insufficient when the environment changes, when the selected behavior fails, or when additional requirements not fully encoded in the reward function become relevant after training.
In these cases, the agent may need to switch to another policy that better matches the current condition, but such switching requires alternative policies to have been generated in advance~\cite{cully2015robots}.
Quality-diversity reinforcement learning (QD-RL) addresses this need by constructing a repertoire of policies that are both high-return and behaviorally diverse, following the broader quality-diversity principle of searching for diverse and high-quality solutions~\cite{cully2018quality}.
In QD-RL, each candidate policy is evaluated by its episodic return and by a rollout-level behavioral descriptor that maps the policy into a behavior space for diversity comparison.
This objective is commonly supported by archive-based illumination, where high-performing elites are retained across behavior descriptor cells~\cite{mouret2015illuminating}, and by local competition, where behaviorally related candidates compete on quality~\cite{lehman2011creatures}.
Thus, the output of QD-RL is not a single best policy, but a repertoire of selectable policies that differ in behavior while maintaining high task performance.

For high-dimensional neural policies, recent QD-RL methods have increasingly introduced learning updates into the archive loop.
Policy-gradient assisted QD refines archive candidates through gradient variation~\cite{nilsson2021pga}, and generalized actor--critic QD-RL studies how modern actor--critic components can be transferred into QD-RL settings~\cite{lim2023understanding}.
Beyond improving return, several methods explicitly use gradient or critic signals to promote behavioral diversity.
Diversity policy gradients move policies toward under-explored behavior regions~\cite{pierrot2022diversity}.
Descriptor-conditioned gradients use learned critics to guide policies toward target behavior descriptors~\cite{faldor2023dcg}, and descriptor-conditioned reinforcement learning further uses the learned actor as a generative model for policies associated with target niches~\cite{faldor2025dcrl}.
Quality-diversity actor--critic methods also use value and successor-feature critics to learn policies that are both high-performing and diverse~\cite{grillotti2024qdac}.
These studies show that learning and critic signals already participate in diversity generation in modern QD-RL.
The remaining gap is therefore not whether gradient or critic information can be used for QD-RL, but what part of the learning process is diversified.
Existing methods mainly use learned value information to improve policy quality, guide descriptor-conditioned generation, or optimize diversity-aware actor objectives, usually within a fixed actor structure.
In contrast, actor structure, critic state, and replay distribution are rarely treated as coupled branch variables that define distinct learning trajectories before archive admission.
This motivates a branch-level view of QD-RL, where a candidate is not only an actor parameter vector but also a structure-conditioned actor--critic learning state.

This branch-level view is important because the behavior of an actor--critic policy is shaped not only by the current actor parameters, but also by the actor function class, the value estimator, and the replay distribution used for policy improvement.
Changing the actor structure changes the effective function class in which a policy is optimized and the state-action dependencies available during learning.
Network structure has already appeared as a meaningful search factor in QD and architecture-search contexts.
QD-based neural architecture search treats architecture as a search object for discovering high-performing networks across different niches~\cite{schneider2022tackling}, and cooperative coevolutionary QD decomposes policy networks into interacting components to improve sample efficiency~\cite{xue2024sample}.
Sparse-training studies further show that dense networks can contain trainable sparse subnetworks~\cite{frankle2019lottery}, and that sparse RL policies can remain competitive while topology choices interact with non-stationary data distributions and value learning~\cite{graesser2022state,tan2023rlx2}.
These results suggest that actor structure is not a neutral implementation detail.
By changing the effective function class and the state-action dependencies available to a policy, structural masks can define different representational subspaces for different branches.

Structural conditioning defines the subspace in which each branch learns, while critic-guided learning determines how the branch moves inside that subspace.
In actor--critic RL, the actor is updated through gradients derived from the learned value function.
Deterministic policy-gradient methods compute actor updates through action-value gradients, and deep actor--critic methods for continuous control build on this mechanism to improve neural policies~\cite{silver2014deterministic,lillicrap2015continuous}.
TD3 further improves the stability of this update process by addressing function-approximation error in actor--critic learning~\cite{fujimoto2018addressing}.
In SV-QD-RL, the critic network is used as a branch-specific guidance signal rather than only as a return estimator.
Because different structure-conditioned branches maintain private critics and replay states, their critics are learned under different state-action distributions and can automatically guide the corresponding actors to explore different local update directions under the same external reward.
This observation motivates coupling structural conditioning with branch-private critic guidance as a mechanism for producing behavioral specialization before archive-level selection.

Based on these observations, we develop structure--value coupled QD-RL (SV-QD-RL), a framework for branch-level diversity search in actor--critic policy repertoires.
Instead of representing each candidate only as an actor parameter vector, SV-QD-RL represents each candidate as a structure-conditioned actor--critic branch.
Each branch contains an actor, a structural mask, a branch-specific critic, a replay state, and evaluation attributes including behavior, return, sparsity, and value profile.
Unlike diversity-objective or descriptor-conditioned critic mechanisms that primarily guide policies toward manually specified diversity rewards or target descriptors, SV-QD-RL ties each branch-specific critic to a structural mask and its matched replay distribution.
The structural mask defines the actor subspace, the replay state determines the data distribution from which the critic is learned, and the private critic generates the value-gradient field that updates the actor inside that subspace.
The archive then evaluates these branches against the distribution already represented by the repertoire.
A candidate is preferred not only when it has high return, but also when its behavior, structural footprint, or critic-value profile moves away from existing archive modes.
In this way, SV-QD-RL complements self-guided actor--critic refinement with archive-level distributional repulsion, making specialization emerge from the interaction between local critic gradients and global repertoire occupancy.

The main contributions of this paper are summarized below.

\begin{enumerate}
	\item We formulate QD-RL as a branch-level diversity search problem over actor parameters, structural conditions, critic states, and replay distributions.
	This formulation extends conventional policy-instance diversification by considering the learning conditions that generate behavioral specialization before archive admission.
	
	\item We propose SV-QD-RL, a structure--value coupled search framework based on structure-conditioned actor--critic branches.
	The framework combines multi-regime neuron-level mask search, branch-private critic guidance, memory-consistent replay refinement, archive-distribution novelty, and archive feedback.  Each branch critic acts as a self-guided value-gradient field for its own masked actor, while archive admission encourages candidates to move away from behavior--structure--value regions that are already represented.
	
	\item We provide empirical and diagnostic analyses of structure-conditioned critic-guided diversification.
	The analyses examine how structural conditioning, critic differentiation, and replay-consistent refinement contribute to behavioral diversity, archive quality, and repertoire selection on continuous-control benchmarks.
\end{enumerate}

We evaluate SV-QD-RL on MuJoCo continuous-control benchmarks from three complementary perspectives.
First, we compare the final archives using standard QD metrics, including QD-score, coverage, and best return.
Second, we analyze whether structural conditioning and differentiated critic guidance make distinct and complementary contributions to behavioral diversity.
Third, we use a fixed-archive deployment evaluation to test whether the learned archive contains selectable policy alternatives under changing behavior-level requirements.
The results show that SV-QD-RL achieves superior archive quality and behaviorally useful diversity across the MuJoCo benchmarks.
The fixed-archive deployment experiment further confirms that the learned repertoire provides practical policy alternatives when behavior-level requirements change.

The remainder of this paper is organized as follows.
Section~\ref{sec:related} reviews related work.
Section~\ref{sec:method} presents the SV-QD-RL framework and optimization procedure.
Section~\ref{sec:experiments} reports the experimental results.
Finally, Section~\ref{sec:discussion} and Section~\ref{sec:conclusion} discuss the implications and conclude the paper.
\section{Related Work}\label{sec:related}

SV-QD-RL is related to three lines of research: behavior-level diversity in reinforcement learning, architecture and sparsity optimization, and representation-aware actor--critic learning. These areas have provided important tools for constructing diverse policies, improving neural policy efficiency, and stabilizing value-guided policy optimization. However, they are often studied as separate components. Existing QD-RL methods usually organize diversity after rollout evaluation in a behavior descriptor space, sparse or architecture-search methods mainly optimize the structure of a single model, and critic-based methods usually improve policy learning without explicitly treating critic state as an archive-level diversity source. SV-QD-RL connects these directions by treating a candidate policy as a structure-conditioned actor--critic branch rather than only as an actor parameter vector. In this view, behavior, structural footprint, critic state, and replay distribution jointly determine how candidate policies are generated and retained.

\subsection{Behavior Diversity in Reinforcement Learning}

Behavior-diversity methods extend reinforcement learning beyond single-policy reward maximization by encouraging exploration, skill separation, or archive-level variation~\cite{zhao2023diversity}. Classical value-based exploration, such as DQN-style exploration, can diversify data collection, but it does not explicitly maintain a deployable set of distinct policies~\cite{mnih2015human}. Quality-diversity optimization directly addresses this limitation by jointly optimizing solution quality and behavioral coverage~\cite{pugh2016quality,mouret2020illumination}, building on novelty search~\cite{lehman2011novelty}. MAP-Elites stores high-performing elites in behavior cells and has been used to construct repertoires for rapid robotic adaptation after damage~\cite{cully2015robots}. Later QD variants improve scalability, archive organization, and variation operators through covariance adaptation, dynamics-aware selection, deep neuroevolution, and scalable behavior-space partitioning~\cite{fontaine2020covariance,lim2022dynamics,colas2020scaling}. CVT-MAP-Elites further shows that archive partitioning can be scaled to higher-dimensional descriptor spaces by using centroidal Voronoi tessellations instead of a dense grid~\cite{vassiliades2018using}, while subsequent studies emphasize the importance of descriptor design and model-based variation for archive illumination~\cite{vassiliades2018benefits,rodriguez2021model}.

A related group of methods uses intrinsic motivation, population search, or hybrid evolutionary-RL optimization to obtain diverse policies. DIAYN learns distinguishable skills through mutual-information maximization~\cite{eysenbach2018diversity}. CEM-RL and ERL combine evolutionary search with gradient-based policy learning~\cite{pourchot2018cem,khadka2018evolution}, and novelty-driven evolution strategies demonstrate that diversity objectives can help avoid deceptive local optima in high-dimensional policy spaces~\cite{conti2018improving}. Recent QD-RL methods further inject policy-gradient or critic-guided improvement into the archive loop. PGA-MAP-Elites refines archive candidates with policy-gradient updates~\cite{nilsson2021pga}; diversity policy gradients and descriptor-conditioned gradients use learned signals to guide policies toward under-explored or target descriptor regions~\cite{pierrot2022diversity,faldor2023dcg}; descriptor-conditioned RL treats the learned actor as a generator of policies for target niches~\cite{faldor2025dcrl}; and QDAC introduces actor--critic mechanisms for learning high-performing and diverse policies~\cite{grillotti2024qdac}. These studies show that modern QD-RL already benefits from gradient and value information. Nevertheless, most methods still define diversity primarily through realized behavior descriptors and usually keep the actor architecture fixed. SV-QD-RL differs by diversifying the learning branch itself: structural masks, branch-specific critics, replay states, behavior descriptors, and value profiles are treated as coupled variables that can generate specialization before archive admission.

\subsection{Neural Architecture Search and Sparse Representation}

Neural architecture search optimizes network topology through discrete, relaxed, or weight-sharing search spaces~\cite{elsken2019neural}. DARTS enables gradient-based optimization over a relaxed supernet~\cite{liu2019darts}, while hardware-aware NAS incorporates latency, memory, or resource constraints into architecture selection~\cite{cai2020once}. These studies show that network structure can strongly affect both accuracy and deployment cost. In parallel, pruning methods reduce redundant computation by estimating parameter saliency or removing unnecessary connections~\cite{lecun1989optimal,hassibi1993second,han2015deep}. Sparse training further shifts the focus from compressing a finished dense model to learning trainable subnetworks during optimization~\cite{frankle2019lottery,rigl}. Supermask-based approaches also indicate that binary masks can encode task-specific computation paths within a shared parameter space~\cite{wortsman2020supermasks}.

Sparsity is especially nontrivial in reinforcement learning because the data distribution is policy-induced and changes throughout training. Parameters that appear unimportant early may become useful later when the policy visits different states, making conventional pruning less reliable in RL settings~\cite{graesser2022state}. Dynamic sparse RL methods, such as RLx2, address part of this issue by adapting topology during training and improving high-sparsity policy learning~\cite{tan2022rlx2}. However, most sparse RL and NAS-style approaches still focus on finding a compact or efficient version of a single policy, or on selecting one architecture family under resource constraints. They do not usually ask whether different structural conditions can be maintained simultaneously as part of a policy repertoire. SV-QD-RL uses structured masks in a different way: the mask is not only a compression device, but also a repertoire-search variable. Different masks define different actor subspaces, which may lead to different behaviors, different deployment footprints, and different value-learning trajectories. This allows compact policies to be retained not merely because they approximate a dense policy, but because they occupy useful behavioral and structural niches in the archive.

\subsection{Value Representation and Critic Diversity}

Representation learning shapes how reinforcement learning agents evaluate experience, generalize across states, and discover reusable behaviors. Successor-feature methods decompose value into latent state features and reward weights~\cite{bcs}, and skill-discovery methods learn representations that make different behaviors distinguishable~\cite{eysenbach2018diversity}. In actor--critic methods, the critic has a more direct role: it provides the value estimates or value gradients that determine how the actor is updated. Deterministic policy-gradient methods compute policy updates through action-value gradients~\cite{silver2014deterministic}; TD3 improves the stability of this process by reducing function-approximation error~\cite{fujimoto2018addressing}; and advantage-based policy-gradient methods provide another family of critic-dependent policy-improvement signals~\cite{schulman2017proximal}. These methods indicate that the critic is not only an evaluator after rollout, but also a local optimization geometry that shapes the direction and stability of actor learning.

Prior work has connected representation diversity to exploration, robustness, transfer, and sample efficiency~\cite{hassabis2017neuroscience,liu2024adaptive}. Population-based RL suggests that maintaining diverse learners can reduce correlated failures~\cite{parker2020effective}, and ensemble-value methods such as Bootstrapped DQN show that randomized value functions can induce temporally extended exploration~\cite{osband2016deep}. In QD-RL, learned critics can also be used to guide policies toward target descriptors or improve diversity-aware policy optimization. However, critic diversity is rarely coupled with structural sparsity, replay-state consistency, and archive admission. SV-QD-RL addresses this gap by assigning branch-specific critics to structure-conditioned actor branches. Because different masks induce different actor subspaces and state-action visitation distributions, their critics are trained under different data distributions and can produce different value geometries. SV-QD-RL then uses value-profile information as part of archive-level novelty, linking critic diversity with structural and behavioral diversity. This coupling makes diversity generation a learning-level process rather than only a post-hoc selection step based on rollout descriptors.

\section{The Structure--Value Coupled QD-RL Framework}
\label{sec:method}

\subsection{Overview}
\label{subsec:overview}

SV-QD-RL is designed to diversify the learning branches that generate candidate policies, rather than only diversifying policy instances after rollout evaluation.
The central object is a structure-conditioned actor--critic branch.
A branch couples five elements: a masked actor, a structural condition, a branch-private critic, a replay-memory state, and archive attributes such as return, behavior descriptor, sparsity, and value profile.
The structural mask specifies the actor subspace in which the branch can represent policies.
The replay state specifies the data distribution on which the critic is trained.
The branch-private critic then defines the local value-gradient field that updates the actor in that masked subspace.
Thus, branch-specific critic networks automatically guide different structure-conditioned branches to explore different local policy-update directions.
The archive supplies the complementary global signal: it summarizes which behavior--structure--value regions have already been represented and gives novelty pressure to candidates that move away from those occupied distributions.

\begin{figure}[ht]
	\centering
	\includegraphics[width=\columnwidth]{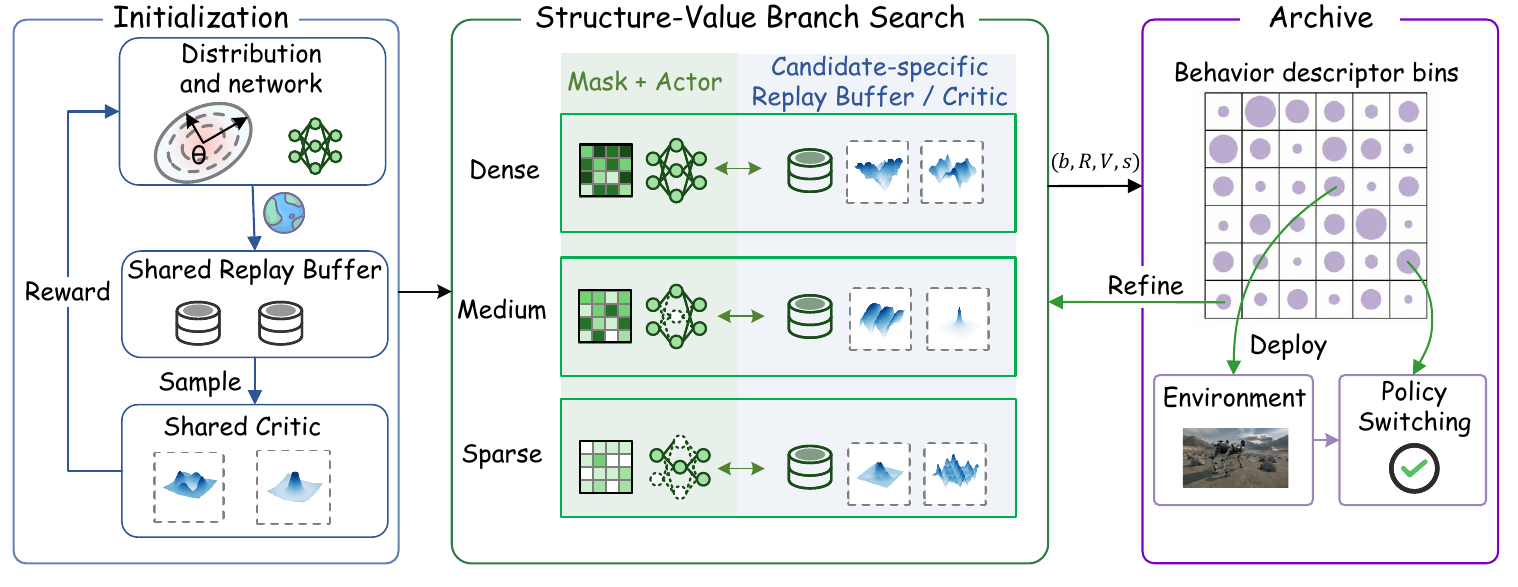}
	\caption{Overall workflow of SV-QD-RL.
		The framework generates structure-conditioned actor--critic branches, lets branch-private critics guide memory-consistent actor refinement, and uses the branch-aware archive to preserve high-quality candidates that move away from already represented behavior--structure--value distributions.}
	\label{fig:method_overview}
\end{figure}

Fig.~\ref{fig:method_overview} summarizes the workflow.
SV-QD-RL starts from a shared actor--critic substrate and creates candidate branches through parameter proposals, structural mask proposals, and branch-private value refinement.
Evaluated branches are admitted to a branch-aware QD archive according to behavior, quality, structural footprint, and value-profile information.
The archive then selects representative branch states for further refinement.
This forms a closed loop: structural proposals create heterogeneous actor subspaces; branch-private critics automatically guide local actor updates; replay memories keep each critic aligned with its branch distribution; and archive-level distributional novelty pushes accepted candidates away from regions that are already populated.
The final output is a policy repertoire, while the main mechanism is self-guided branch diversification coupled with archive-level repulsion from existing behavior--structure--value distributions.

\subsection{Structure-Conditioned Actor--Critic Branches}
\label{subsec:branches}

Consider an MDP 
$\mathcal{M}=(\mathcal{S},\mathcal{A},P,r,\gamma)$, where $\mathcal{S}$ is the state space, $\mathcal{A}$ is the action space, $P$ is the transition kernel, $r$ is the reward function, and $\gamma\in[0,1)$ is the discount factor.
The basic search object in SV-QD-RL is a branch
\begin{equation}
	\chi_i=(\omega_i,\xi_i),
	\label{eq:branch}
\end{equation}
where the learning state is
\begin{equation}
	\omega_i=
	\left(
	\theta_i,\mathbf{m}_i,\phi_i,\bar{\phi}_i,\eta_i,\mathcal{D}_{\eta_i}
	\right),
	\label{eq:branch_state}
\end{equation}
and the evaluation attributes are
\begin{equation}
	\xi_i=
	\left(
	\mathbf{z}_i,\widehat{J}_i,\kappa_i,\mathbf{q}_i
	\right).
	\label{eq:branch_attributes}
\end{equation}
Here, $\theta_i$ denotes the actor parameters, $\mathbf{m}_i$ the structural mask, $\phi_i$ and $\bar{\phi}_i$ the critic and target-critic parameters, $\eta_i$ the replay-memory identifier, and $\mathcal{D}_{\eta_i}$ the replay state.
The replay state is an explicit part of the branch because off-policy value learning should use data distributions consistent with the actor structure and critic state of the same branch.
The attributes $\mathbf{z}_i$, $\widehat{J}_i$, $\kappa_i$, and $\mathbf{q}_i$ denote the behavioral descriptor, evaluated return, structural sparsity, and critic-induced value profile.

We use neuron-level binary masks to impose structural conditions on actor networks.
Let $\pi_{\theta_i}$ be an actor with $L-1$ hidden layers and one output layer.
For hidden layer $l$, let $d_l$ be its width, $\mathbf{W}_i^{(l)}\in\mathbb{R}^{d_l\times d_{l-1}}$ and $\mathbf{b}_i^{(l)}\in\mathbb{R}^{d_l}$ be its weight matrix and bias vector, and $\mathbf{m}_i^{(l)}\in\{0,1\}^{d_l}$ be its mask.
The full mask is
\begin{equation}
	\mathbf{m}_i=
	\left(
	\mathbf{m}_i^{(1)},\ldots,\mathbf{m}_i^{(L-1)}
	\right).
	\label{eq:mask}
\end{equation}
For state $s\in\mathcal{S}$, the masked forward propagation is
\begin{equation}
	\mathbf{h}_i^{(0)}(s)=s,
	\label{eq:masked_input}
\end{equation}
\begin{equation}
\begin{aligned}
	\mathbf{h}_i^{(l)}(s)
	&=
	\mathbf{m}_i^{(l)}
	\odot
	\sigma\!\left(
	\mathbf{W}_i^{(l)}\mathbf{h}_i^{(l-1)}(s)
	+
	\mathbf{b}_i^{(l)}
	\right),\\
	&\hspace{5.5em} l=1,\ldots,L-1,
\end{aligned}
\label{eq:masked_hidden}
\end{equation}
and
\begin{equation}
	\pi_{\theta_i,\mathbf{m}_i}(s)
	=
	g\!\left(
	\mathbf{W}_i^{(L)}\mathbf{h}_i^{(L-1)}(s)
	+
	\mathbf{b}_i^{(L)}
	\right).
	\label{eq:masked_output}
\end{equation}
The output layer is kept unmasked, so structural variation acts through hidden representation bottlenecks rather than by removing action dimensions.

The sparsity level is computed from the parameter count after structured channel pruning.
Let $d_0$ and $d_L$ denote the input and output dimensions, and let $r_i^{(l)}=\|\mathbf{m}_i^{(l)}\|_0$ be the number of retained hidden units in layer $l$.
The dense actor parameter count is
\begin{equation}
	P_{\mathrm{dense}}
	=
	\sum_{l=1}^{L}
	\left(
	d_l d_{l-1}+d_l
	\right),
	\label{eq:dense_param_count}
\end{equation}
and the structurally pruned actor parameter count is
\begin{equation}
	\begin{aligned}
		P_i
		={}&
		d_0 r_i^{(1)}+r_i^{(1)}
		+
		\sum_{l=2}^{L-1}
		\left(
		r_i^{(l-1)}r_i^{(l)}
		+
		r_i^{(l)}
		\right)\\
		&+
		r_i^{(L-1)}d_L+d_L .
	\end{aligned}
	\label{eq:sparse_param_count}
\end{equation}
The structural sparsity is
\begin{equation}
	\kappa_i=
	1-\frac{P_i}{P_{\mathrm{dense}}}.
	\label{eq:sparsity}
\end{equation}
This definition is consistent with the implementation, where neuron masks are expanded into weight masks and the output layer is protected.
Given two branches, their structural-footprint distance is
\begin{equation}
	d_{\mathrm{str}}(i,j)=|\kappa_i-\kappa_j|.
	\label{eq:structural_distance}
\end{equation}

Behavioral attributes are computed from evaluation rollouts.
Let $\{\tau_i^{(e)}\}_{e=1}^{n_e}$ be trajectories generated by $\pi_{\theta_i,\mathbf{m}_i}$.
The behavioral descriptor and evaluated return are
\begin{equation}
	\mathbf{z}_i=
	\psi\!\left(
	\{\tau_i^{(e)}\}_{e=1}^{n_e}
	\right),
	\label{eq:behavior_descriptor}
\end{equation}
and
\begin{equation}
	\widehat{J}_i
	=
	\frac{1}{n_e}
	\sum_{e=1}^{n_e}
	\sum_{t=0}^{T_e-1}
	\gamma^t
	r(s_t^{(e)},a_t^{(e)}),
	\label{eq:evaluated_return}
\end{equation}
where $\psi(\cdot)$ is the task-dependent descriptor mapping.
The behavioral distance is
\begin{equation}
	d_{\mathrm{beh}}(i,j)=\|\mathbf{z}_i-\mathbf{z}_j\|_2 .
	\label{eq:behavior_distance}
\end{equation}

Value-level attributes are computed from the branch-specific critic on a common reference batch.
Let $\mathcal{B}_Q=\{(\bar{s}_b,\bar{a}_b)\}_{b=1}^{B_Q}$ be a batch of state-action pairs sampled from a reference buffer shared across branch comparisons.
For a TD3-style twin critic, the scalar value used for comparison is
\begin{equation}
	q_{i,b}
	=
	\min_{h\in\{1,2\}}
	Q_{\phi_{i,h}}(\bar{s}_b,\bar{a}_b),
	\quad b=1,\ldots,B_Q .
	\label{eq:value_profile}
\end{equation}
The value profile is $\mathbf{q}_i=(q_{i,1},\ldots,q_{i,B_Q})$.
To compare critics independently of scale, profiles are converted into smoothed empirical histograms over common bins $\{I_r\}_{r=1}^{R_Q}$,
\begin{equation}
	p_i(r)=
	\frac{
		\sum_{b=1}^{B_Q}\mathbb{I}[q_{i,b}\in I_r]+\epsilon_Q
	}{
		B_Q+R_Q\epsilon_Q
	},
	\quad r=1,\ldots,R_Q .
	\label{eq:value_distribution}
\end{equation}
The value-profile distance is the symmetric KL divergence
\begin{equation}
	d_{\mathrm{val}}(i,j)
	=
	\frac{1}{2}
	\left[
	\mathrm{KL}(\mathbf{p}_i\|\mathbf{p}_j)
	+
	\mathrm{KL}(\mathbf{p}_j\|\mathbf{p}_i)
	\right].
	\label{eq:value_distance}
\end{equation}

The representation in Eq.~\eqref{eq:branch} defines three coupled axes used by SV-QD-RL.
The descriptor $\mathbf{z}_i$ identifies realized behavior, $\kappa_i$ specifies the structural condition, and $\mathbf{q}_i$ characterizes the value geometry experienced by the branch.
The branch critic connects these axes to learning dynamics.
For a branch $i$, define the critic-induced action-gradient field
\begin{equation}
\mathcal{G}_i(s)
=
\nabla_a Q_{\phi_i}(s,a)\big|_{a=\pi_{\theta_i,\mathbf{m}_i}(s)} .
\label{eq:critic_guidance_field}
\end{equation}
This field is learned from the replay distribution associated with the same masked actor.
Consequently, two branches can receive different update directions even when they are evaluated under the same reward function and have nearby rollout descriptors.
SV-QD-RL uses the value profile $\mathbf{q}_i$ to make this otherwise hidden critic-induced geometry visible to archive admission and continuation selection.
The archive compares $\mathbf{q}_i$ with the value-profile distribution of retained branches, so a branch can be favored when its critic represents a different local value geometry even before this difference is fully expressed as a large behavior displacement.
Thus, behavioral specialization is encouraged not by manually assigning separate rewards, but by allowing branch-private critics to self-guide actor refinement while the archive automatically repels new candidates from already represented behavior--structure--value modes.

\subsection{Structure--Value Branch Search}
\label{subsec:branch_search}

SV-QD-RL generates candidate branches through a structure--value search loop.
The search state contains an actor proposal distribution, a mask proposal distribution, a shared actor--critic substrate, and a pool of retained branch-private states.
The actor distribution proposes parameter-space candidates, the mask distribution proposes structural conditions, and the retained pool provides branch states for continued local refinement.

At initialization, the actor proposal distribution is set to $(\boldsymbol{\mu}_{\theta}^{(0)},\mathbf{c}_{\theta}^{(0)})$, where $\mathbf{c}_{\theta}^{(0)}$ is a diagonal covariance vector.
The mask proposal distribution is initialized as $(\boldsymbol{\rho}^{(0)},\mathbf{c}_{m}^{(0)})$, where $\boldsymbol{\rho}^{(0)}\in[0,1]^{d_m}$ and $d_m=\sum_{l=1}^{L-1}d_l$.
When a neuron-importance prior is available, $\boldsymbol{\rho}^{(0)}$ is biased by normalized weight magnitude, inverse actor-proposal covariance, and the target sparsity range.
We denote this compactly as
\begin{equation}
	\boldsymbol{\rho}^{(0)}
	=
	\mathcal{G}_{\mathrm{init}}
	\left(
	\widetilde{|W|},
	\widetilde{\mathbf{c}_{\theta}^{-1}},
	\kappa_{\min},
	\kappa_{\max}
	\right).
	\label{eq:mask_init}
\end{equation}
If the prior is unavailable, uniform initialization is used before applying the same target-sparsity bias.
The shared critic state $\phi_{\mathrm{base}}^{(0)}$ and replay state $\mathcal{D}_{\mathrm{base}}^{(0)}$ provide the initial substrate for branch generation.

At each global iteration, candidate branches are produced from three sources.
First, a CEM-style parameter proposal samples actor candidates from the current actor distribution.
These candidates are evaluated and inserted into the archive when admissible.
Second, the archive selects representative branches for memory-consistent actor--critic refinement.
Each selected branch is refined using the replay memory associated with its structural identity.
Third, the mask proposal module generates new structural candidates.
Within one iteration, candidate sets are processed in the order of parameter proposals, refined archive representatives, and structural mask proposals.

For a mask candidate, a target sparsity level $\bar{\kappa}\in\mathcal{K}$ is selected, where $\mathcal{K}=\{\bar{\kappa}_1,\ldots,\bar{\kappa}_Q\}$.
Continuous neuron-mask scores are sampled from the mask proposal distribution, adjusted according to the selected sparsity regime, and thresholded by $\tau_m$ to obtain binary neuron masks.
The flattened mask is split into layer-wise masks and expanded into weight masks, while the output layer remains unpruned.
The resulting masked actors are evaluated, assigned behavior descriptors, sparsity values, critic states, value profiles, and memory identifiers, and then submitted to the archive.

The role of actor--critic learning in this loop is local self-guided branch refinement rather than replacement of QD search.
Given a selected branch $\chi_i$ with learning state $\omega_i^{(t)}$, memory-consistent refinement is written as
\begin{equation}
	\omega_i^{(t,+)}
	=
	\mathcal{U}_{\mathrm{refine}}
	\left(
	\omega_i^{(t)},
	\mathcal{D}_{\eta_i}^{(t)}
	\right),
	\quad
	\chi_i\in \mathcal{R}_{\mathrm{sel}}^{(t)},
	\label{eq:refinement}
\end{equation}
where $\mathcal{R}_{\mathrm{sel}}^{(t)}$ is the archive-selected branch subset.
In the implementation, $\mathcal{U}_{\mathrm{refine}}$ is instantiated by TD3.
For branch $i$, the deterministic actor update follows the local critic gradient
\begin{equation}
\begin{aligned}
    \nabla_{\theta_i} \mathcal{J}_i
    \approx
    \frac{1}{|\mathcal{B}_i|}
    \sum_{s\in\mathcal{B}_i}
    \nabla_a Q_{\phi_i}(s,a)\big|_{a=\pi_{\theta_i,\mathbf{m}_i}(s)}
    \nabla_{\theta_i}\pi_{\theta_i,\mathbf{m}_i}(s),
\end{aligned}
\label{eq:branch_private_policy_gradient}
\end{equation}
where $\mathcal{B}_i\subset\mathcal{D}_{\eta_i}$ is sampled from the replay memory matched to the branch.
Equation~\eqref{eq:branch_private_policy_gradient} makes the self-guided role of the critic explicit: the actor is updated by the value-gradient field of its own branch critic, not by a shared critic or an externally specified diversity target.
Different critics can therefore produce different local ascent directions under the same reward, allowing the archive to obtain behaviorally distinct policies through branch-local learning dynamics.
The update is applied only after matching the branch to its replay memory, which keeps the mask, critic, and data distribution aligned during off-policy refinement.

The actor proposal distribution is updated from parameter-space elites using a CEM-style update,
\begin{equation}
	(\boldsymbol{\mu}_{\theta}^{(t+1)},\mathbf{c}_{\theta}^{(t+1)})
	=
	\mathcal{U}_{\mathrm{CEM}}^{\theta}
	\left(
	\boldsymbol{\mu}_{\theta}^{(t)},
	\mathbf{c}_{\theta}^{(t)},
	\mathcal{E}_{\theta}^{(t)}
	\right),
	\label{eq:actor_update}
\end{equation}
where $\mathcal{E}_{\theta}^{(t)}$ denotes the parameter-space elite set.
The mask proposal distribution is updated by CEM using high-scoring structural candidates.
Let $\mathcal{E}_m^{(t)}$ denote the mask elite set selected according to a proposal score that combines evaluated return and structural preference.
The mask distribution is updated as
\begin{equation}
	(\boldsymbol{\rho}^{(t+1)},\mathbf{c}_m^{(t+1)})
	=
	\mathcal{U}_{\mathrm{CEM}}^m
	\left(
	\boldsymbol{\rho}^{(t)},
	\mathbf{c}_m^{(t)},
	\mathcal{E}_m^{(t)}
	\right).
	\label{eq:mask_update}
\end{equation}
The updated substrate provides initialization for future branches, while archive-retained branches preserve their private actor--critic states.

\subsection{Memory-Consistent Branch Refinement}
\label{subsec:memory_refinement}

SV-QD-RL treats replay memory as part of the branch state.
Branches with different masks may induce different state-action visitation distributions, and refining all branches from a single global replay memory can weaken the intended structure--value coupling.
The archive therefore maintains an explicit association between each stored branch and its replay-memory identity.

For local refinement, representative archive entries are selected by nearest-better clustering over archive descriptors and quality.
In the implementation, the archive is partitioned by sparsity thresholds $[0.10,0.60,0.85]$, a high-sparsity protection rule is applied, and at most $N_{\mathrm{ft}}$ selected branches are refined in one iteration.
For each selected branch, the corresponding actor, mask, critic, target critic, sparsity value, and memory identifier are retrieved together.
Sparse branches retrieve their mask-indexed replay memories.
Dense branches use private dense-local memories initialized from the global dense memory.
The selected branch is refined under its matched memory, re-evaluated, and reinserted into the archive with its updated actor, critic, behavior descriptor, value profile, sparsity value, and memory identifier.

This mechanism makes the archive an active training pool rather than a passive collection of evaluated policies.
The archive selects branch-private learning states, matched memories preserve the replay distributions from which critics are learned, and branch-specific critics provide the local value-gradient geometry for actor improvement.
Thus, memory-consistent refinement implements the structure--value--memory coupling required for self-guided branch-level diversity search.

\subsection{Branch-Aware QD Archive}
\label{subsec:branch_archive}

The archive converts evaluated branches into a stable policy repertoire and selects branch-private states for continued search.
Unlike a behavior-only archive, the branch-aware archive considers behavioral novelty, structural condition, and value-profile divergence.
Let $\mathcal{R}^{(t)}$ denote the archive at iteration $t$.
For a branch $\chi_i$, its behavior cell and structural tier are
\begin{equation}
	b_i=\mathcal{B}(\mathbf{z}_i),
	\quad
	g_i=\mathcal{T}(\kappa_i),
	\label{eq:cell_tier}
\end{equation}
where $\mathcal{B}(\cdot)$ maps a descriptor to a behavior cell and $\mathcal{T}(\cdot)$ maps a sparsity value to a structural tier.
The archive is organized as
\begin{equation}
	\mathcal{R}^{(t)}
	=
	\bigcup_{b,g}
	\mathcal{R}_{b,g}^{(t)} .
	\label{eq:archive}
\end{equation}

For archive comparison, SV-QD-RL uses an effective branch score that preserves raw evaluated return for reporting while allowing controlled structural and value-profile balancing during admission,
\begin{equation}
	\begin{aligned}
		\widehat{J}_{\mathrm{eff}}(\chi_i)
		={}&
		\widehat{J}_i
		+
		B_{\mathrm{tier}}(\chi_i)\\
		&+
		\mathbb{I}[\kappa_i\ge \tau_{\kappa}]
		|\widehat{J}_i|
		\left(
		\omega_1\kappa_i+
		\omega_2\min(\kappa_i,\kappa_{\max})
		\right).
	\end{aligned}
	\label{eq:effective_score}
\end{equation}
The tier bonus is activated only for under-represented structural tiers,
\begin{equation}
	B_{\mathrm{tier}}(\chi_i)
	=
	|\widehat{J}_i|\,
	\omega_3
	\min
	\left(
	\frac{|n_{g_i}-q_{g_i}|}{q_{g_i}},
	\rho_{\max}
	\right),
	\quad \omega_3>0,
	\label{eq:tier_bonus}
\end{equation}
where $n_{g_i}$ and $q_{g_i}$ are the current count and minimum quota of tier $g_i$.
The effective score is used only for admission and replacement, and all reported returns use $\widehat{J}_i$.

For a candidate branch $\chi_c$, let
\begin{equation}
	n(c)
	=
	\arg\min_{\chi_i\in\mathcal{R}^{(t)}}
	d_{\mathrm{beh}}(c,i),
	\quad
	\delta_{\mathrm{beh}}(c)
	=
	d_{\mathrm{beh}}(c,n(c))
	\label{eq:nearest_archive}
\end{equation}
be its nearest archive neighbor in behavior space.
The behavioral novelty threshold is adaptive,
\begin{equation}
	\tau_{\mathrm{beh}}^{(t)}(c)
	=
	\frac{\bar{d}_{\mathrm{NN}}^{(t)}}{\lambda_{\mathrm{th}}}
	r_{\kappa}(\kappa_c)
	r_{\mathrm{cap}}(|\mathcal{R}^{(t)}|)
	r_{\mathrm{tier}}(g_c),
	\label{eq:adaptive_behavior_threshold}
\end{equation}
where $\bar{d}_{\mathrm{NN}}^{(t)}$ is the average archive nearest-neighbor distance and $\lambda_{\mathrm{th}}$ is the threshold divisor.
The factors $r_{\kappa}$, $r_{\mathrm{cap}}$, and $r_{\mathrm{tier}}$ are positive relaxation factors determined by candidate sparsity, archive fill level, and structural-tier occupancy, respectively.

Value-profile novelty contributes to the candidate score before the quality-replacement condition is evaluated.
This term is not a hand-crafted behavior target; it gives archive-level visibility to critic-induced learning geometries that may not yet appear as large descriptor differences.
More generally, the archive defines a distributional repulsion signal over the already retained branches.
For a candidate $u$ and archive $\mathcal{R}$, we define
\begin{equation}
d_{\mathrm{arc}}(u,\mathcal{R})
=
\min_{j\in\mathcal{R}}
\left[
\lambda_b \tilde d_b(u,j)
+
\lambda_s \tilde d_{\mathrm{str}}(u,j)
+
\lambda_v \tilde d_{\mathrm{val}}(u,j)
\right],
\label{eq:archive_distribution_distance}
\end{equation}
where the three distances are normalized behavior, structure, and value-profile distances.
A large $d_{\mathrm{arc}}$ means that the candidate is far from the nearest occupied archive mode in the joint behavior--structure--value space.
The value bonus is defined as
\begin{equation}
\begin{aligned}
	B_{\mathrm{val}}(\chi_c)
	&=
	\mathbb{I}
	\left[
	\tau_{\mathrm{val}}^{\mathrm{bonus}}
	<
	d_{\mathrm{val}}(c,n(c))
	\le
	\tau_{\mathrm{val}}^{\mathrm{add}}
	\right]\\
	&\quad{}
	\omega_{\mathrm{val}}
	|\widehat{J}_c|
	\frac{d_{\mathrm{val}}(c,n(c))}
	{\tau_{\mathrm{val}}^{\mathrm{add}}}.
\end{aligned}
\label{eq:value_bonus}
\end{equation}

In this formulation, $b_{\mathrm{val}}$ is the value-profile component of archive-distance repulsion.
The archive therefore behaves as an empirical distribution of discovered learning states, and a new branch is automatically encouraged to move away from the nearest existing mode rather than duplicate an already represented critic geometry.

The final comparison score is
\begin{equation}
	S(\chi_c)
	=
	\widehat{J}_{\mathrm{eff}}(\chi_c)
	+
	B_{\mathrm{val}}(\chi_c).
	\label{eq:final_score}
\end{equation}
For archive entries, $S(\chi_i)$ denotes their stored comparison score.

The candidate is admitted if it improves at least one branch-aware criterion,
\begin{equation}
	\begin{aligned}
		\operatorname{Adm}(\chi_c)=1
		\quad \text{if}\quad
		&
		\delta_{\mathrm{beh}}(c)>\tau_{\mathrm{beh}}^{(t)}(c)\\
		&\text{or}\quad
		d_{\mathrm{str}}(c,n(c))\ge \tau_{\mathrm{str}}\\
		&\text{or}\quad
		d_{\mathrm{val}}(c,n(c))>\tau_{\mathrm{val}}^{\mathrm{add}}\\
		&\text{or}\quad
		S(\chi_c)>S(n(c)).
	\end{aligned}
	\label{eq:admission}
\end{equation}
A dense branch with $\kappa_c<\tau_{\kappa}$ is additionally constrained by a dense-ratio cap,
\begin{equation}
	\frac{
		|\{\chi_i\in\mathcal{R}^{(t)}:\kappa_i<\tau_{\kappa}\}|
	}{
		|\mathcal{R}^{(t)}|
	}
	\le
	\rho_{\mathrm{dense}}^{\max}.
	\label{eq:dense_cap}
\end{equation}
This cap prevents the archive from being dominated by dense branches when sparse structural tiers are under-represented.
When the archive exceeds capacity, pruning first enforces the dense cap, then protects under-represented structural tiers, and finally removes low-quality or low-diversity branches.

After admission, the archive selects branch states for continued refinement using nearest-better clustering (NBC) over archive descriptors and quality.
NBC is used because continuation in SV-QD-RL is a local-improvement problem rather than a purely geometric partitioning problem.
Each branch should either move toward a nearby better solution or become the root of a separate basin when no suitable better neighbor exists.
This makes NBC naturally reward-aware and avoids predefining the number of clusters.
Compared with $k$-means, density clustering, or agglomerative clustering, NBC identifies the behavior--return basins in which branch-private critics should continue self-guided refinement.
It therefore allocates continuation budget to high-quality representatives from distinct local basins instead of repeatedly refining redundant elites.
The archive is first partitioned into structural tiers,
\begin{equation}
	\mathcal{G}_g
	=
	\{
	\chi_i\in\mathcal{R}^{(t+1)}
	:
	\alpha_g\le \kappa_i < \alpha_{g+1}
	\}.
	\label{eq:tiers}
\end{equation}
Within each tier, the better-neighbor set of branch $\chi_i$ is
\begin{equation}
	\mathcal{N}_{i}^{+,(g)}
	=
	\{
	\chi_j\in\mathcal{G}_g:
	\widehat{J}_j>\widehat{J}_i
	\}.
	\label{eq:better_neighbor_set}
\end{equation}
If $\mathcal{N}_{i}^{+,(g)}\neq\emptyset$, its nearest better neighbor is
\begin{equation}
	\operatorname{nb}(i)
	=
	\arg\min_{\chi_j\in\mathcal{N}_{i}^{+,(g)}}
	d_{\mathrm{beh}}(i,j).
	\label{eq:nearest_better}
\end{equation}
Branches without a better neighbor, or with a nearest-better link longer than $\tau_{\mathrm{NBC}}$, become cluster roots.
The threshold $\tau_{\mathrm{NBC}}$ is determined by a percentile of nearest-better distances.
The remaining links partition each tier into clusters $\{C_r^{(g)}\}_{r=1}^{R_g}$.
For each cluster, the representative is
\begin{equation}
	\operatorname{rep}(C_r^{(g)})
	=
	\arg\max_{\chi_i\in C_r^{(g)}} \widehat{J}_i .
	\label{eq:representative}
\end{equation}
The retained branch pool is
\begin{equation}
	\mathcal{H}^{(t+1)}
	=
	\operatorname{Select}_{N_H}
	\left(
	\bigcup_g
	\bigcup_{r=1}^{R_g}
	\{
	\operatorname{rep}(C_r^{(g)})
	\}
	\cup
	\mathcal{P}_{\mathrm{hi}}
	\right),
	\label{eq:retained_pool}
\end{equation}
where $\mathcal{P}_{\mathrm{hi}}$ denotes protected high-sparsity representatives and $N_H$ is the continuation budget.
Thus, the archive determines which behaviorally useful and structurally distinct branch states are carried forward.

\begin{figure}[!t]
	\centering
	\includegraphics[width=\columnwidth]{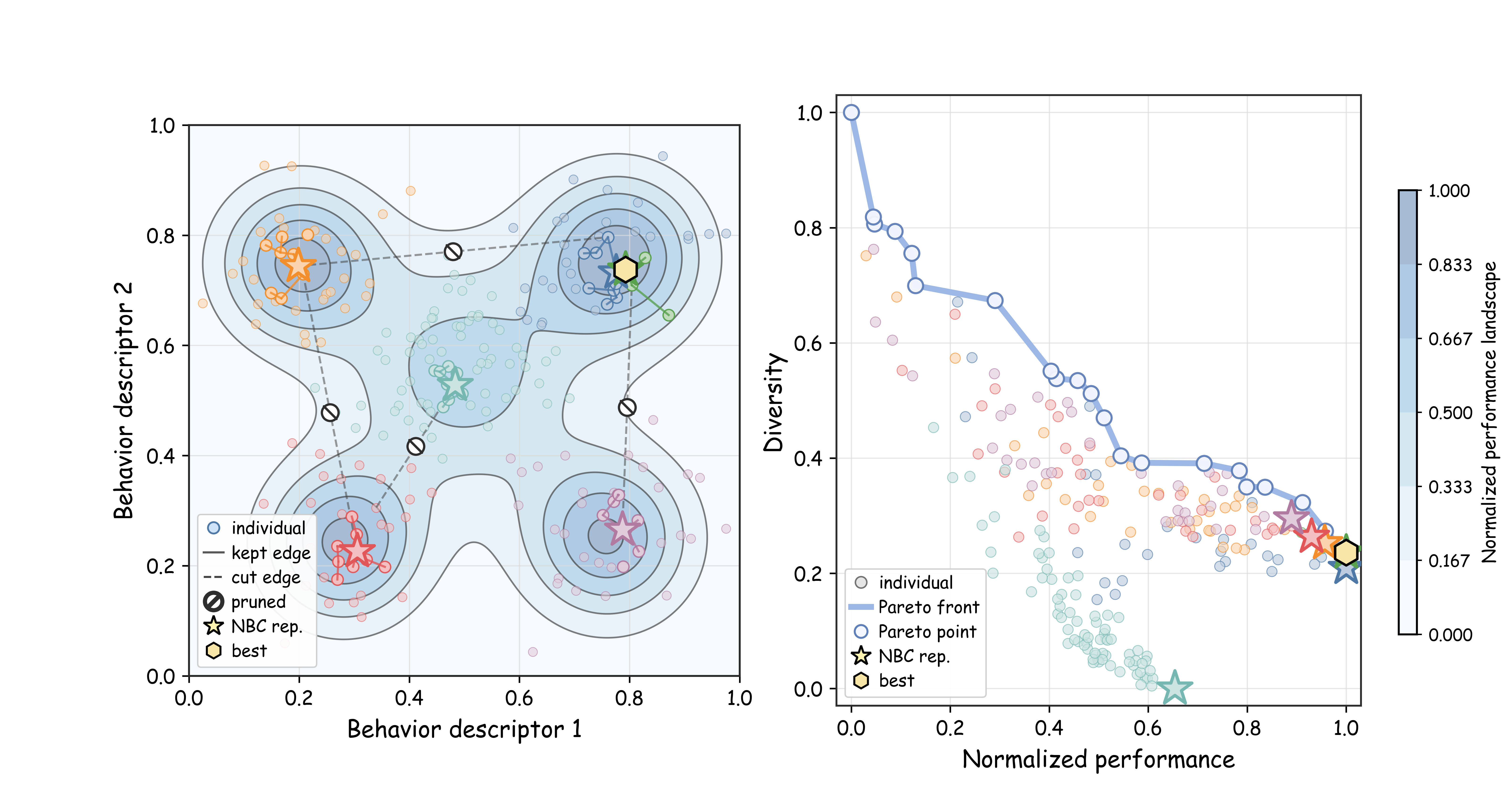}
	\caption{Nearest-better clustering for branch-state selection.
		Branches are embedded in behavior space and linked to their nearest higher-return neighbors.
		Removing overly long links yields local behavioral clusters, from which representative branch-private states are selected for continued refinement.}
	\label{fig:parameter_learning}
\end{figure}

\subsection{Optimization Procedure}
\label{subsec:optimization}

Algorithm~\ref{alg:overall_optimization} summarizes the complete procedure.
Each global iteration contains parameter proposal, self-guided branch refinement, structural mask proposal, archive admission, and continuation-state selection.
The final archive contains policies together with behavior descriptors, masks, sparsity values, critics, target critics, replay-memory identifiers, and value profiles.

\begin{algorithm}[t]
	\caption{SV-QD-RL Optimization Procedure}
	\label{alg:overall_optimization}
	\begin{algorithmic}[1]
		\footnotesize
		\AlgInput{Environment $\mathcal{M}$, archive $\mathcal{R}$, dense memory $\mathcal{D}_d$, sparse memory set $\{\mathcal{D}_s^k\}$, iteration budget $M$}
		\AlgOutput{Branch-aware QD archive $\mathcal{R}$}
		\AlgPhase{Initialization}
		\State Initialize actor proposal distribution, mask proposal distribution, shared actor--critic substrate, archive $\mathcal{R}$, dense memory $\mathcal{D}_d$, and sparse memory containers
		\AlgPhase{Structure--value branch search}
		\For{$t=0$ to $M-1$}
		\State Generate parameter-space candidates using a CEM-style proposal operator
		\State Evaluate candidates and insert admissible branches into $\mathcal{R}$
		\State Select representative archive branches using nearest-better clustering over archive descriptors and quality
		\For{each selected branch}
		\State Retrieve actor, mask, critic, target critic, sparsity value, and memory identifier
		\If{the branch is sparse}
		\State Retrieve its mask-indexed replay memory $\mathcal{D}_s^k$
		\Else
		\State Create a private dense-local memory initialized from $\mathcal{D}_d$
		\EndIf
		\State Refine the branch using its private critic on the matched replay memory
		\State Re-evaluate and reinsert the refined branch into $\mathcal{R}$
		\EndFor
		\State Generate structural mask candidates from the mask proposal distribution
		\State Evaluate mask candidates and compute behavior descriptors, sparsity values, value profiles, and memory identifiers
		\State Insert admissible mask candidates into $\mathcal{R}$
		\State Update archive capacity, synchronize memory identifiers, and compute QD metrics
		\State Update actor and mask proposal distributions using selected branch information
		\EndFor
		\State \Return $\mathcal{R}$
	\end{algorithmic}
\end{algorithm}

The procedure forms a closed branch-generation loop.
Parameter proposals expand the policy search frontier, memory-consistent refinement lets branch-private critics improve selected actors under matched replay distributions, mask proposals expand the structural frontier, and the branch-aware archive determines which behaviorally useful learning states are retained.
Consequently, the final archive is not a collection of independently evaluated policies, but a policy repertoire produced by repeated interaction among structural conditioning, self-guided critics, replay memories, archive-level distributional repulsion, and QD selection.

\section{Experiments}
\label{sec:experiments}

\begin{table*}[!t]
\centering
\caption{Archive performance under the unified speed--contact projection. Values are mean $\pm$ standard deviation over five seeds; best results are bold and second-best results are underlined.}
\label{tab:main_results}
\scriptsize
\setlength{\tabcolsep}{1.8pt}
\renewcommand{\arraystretch}{0.92}
\resizebox{0.88\textwidth}{!}{%
\begin{tabular}{llccccc}
\toprule
Env. & Algorithm & QD-Score & Cov. (\%) & Eff. Cells & Best Return & Mean Elite \\
\midrule
\multirow{8}{*}{Hopper}
& MAP-Elites & $4.05{\times}10^4 \pm 1.12{\times}10^3$ & $2.88\pm0.16$ & $72\pm4$ & $1117.0\pm11.87$ & $963.7\pm24.6$ \\
& NSLC & $6.60{\times}10^4 \pm 1.74{\times}10^3$ & $5.28\pm0.21$ & $132\pm5$ & $1095.7\pm4.68$ & $949.5\pm18.3$ \\
& PGA-MAP-Elites & $1.48{\times}10^5 \pm 5.36{\times}10^3$ & $7.40\pm0.32$ & $185\pm8$ & $\underline{3407.7\pm56.18}$ & $404.5\pm17.8$ \\
& CMA-MAEGA & $1.77{\times}10^4 \pm 9.64{\times}10^2$ & $3.16\pm0.18$ & $79\pm5$ & $1073.0\pm10.68$ & $212.6\pm11.5$ \\
& DNS & $3.17{\times}10^4 \pm 1.21{\times}10^3$ & $2.80\pm0.14$ & $70\pm4$ & $1064.3\pm5.41$ & $409.5\pm13.9$ \\
& QDAC & $\underline{2.03{\times}10^5 \pm 7.82{\times}10^3}$ & $\underline{13.88\pm0.48}$ & $\underline{347\pm12}$ & $332.7\pm12.33$ & $86.0\pm6.7$ \\
& EDOCS & $4.42{\times}10^4 \pm 1.58{\times}10^3$ & $1.00\pm0.08$ & $25\pm2$ & $1831.3\pm4.89$ & $\underline{1554.6\pm28.5}$ \\
& \textbf{SV-QD-RL} & $\mathbf{3.62{\times}10^5 \pm 8.95{\times}10^3}$ & $\mathbf{16.24\pm0.36}$ & $\mathbf{406\pm9}$ & $\mathbf{3633.8\pm9.06}$ & $\mathbf{1718.6\pm25.7}$ \\
\midrule
\multirow{8}{*}{HalfCheetah}
& MAP-Elites & $2.30{\times}10^5 \pm 6.84{\times}10^3$ & $4.44\pm0.22$ & $111\pm6$ & $3852.6\pm25.09$ & $2430.8\pm52.4$ \\
& NSLC & $2.18{\times}10^5 \pm 5.91{\times}10^3$ & $6.44\pm0.27$ & $161\pm7$ & $3271.1\pm15.20$ & $1864.1\pm41.8$ \\
& PGA-MAP-Elites & $1.95{\times}10^5 \pm 8.46{\times}10^3$ & $\underline{23.24\pm0.91}$ & $\underline{581\pm23}$ & $2574.1\pm141.47$ & $303.5\pm24.7$ \\
& CMA-MAEGA & $3.48{\times}10^5 \pm 9.75{\times}10^3$ & $5.52\pm0.24$ & $138\pm6$ & $\underline{4897.4\pm23.30}$ & $\underline{2550.8\pm47.5}$ \\
& DNS & $9.22{\times}10^4 \pm 3.86{\times}10^3$ & $3.56\pm0.19$ & $89\pm5$ & $1927.6\pm15.63$ & $1021.9\pm35.6$ \\
& QDAC & $3.81{\times}10^5 \pm 1.46{\times}10^4$ & $21.84\pm0.76$ & $546\pm19$ & $2653.0\pm122.76$ & $197.8\pm16.2$ \\
& EDOCS & $\underline{8.48{\times}10^5 \pm 2.28{\times}10^4}$ & $17.16\pm0.58$ & $429\pm15$ & $2561.7\pm9.52$ & $1875.7\pm39.4$ \\
& \textbf{SV-QD-RL} & $\mathbf{2.04{\times}10^6 \pm 4.72{\times}10^4}$ & $\mathbf{25.36\pm0.62}$ & $\mathbf{634\pm16}$ & $\mathbf{13325.8\pm30.98}$ & $\mathbf{5835.4\pm86.7}$ \\
\midrule
\multirow{8}{*}{Walker2d}
& MAP-Elites & $7.91{\times}10^4 \pm 2.44{\times}10^3$ & $4.36\pm0.20$ & $109\pm5$ & $1337.5\pm13.03$ & $952.2\pm28.6$ \\
& NSLC & $6.50{\times}10^4 \pm 2.37{\times}10^3$ & $4.96\pm0.23$ & $124\pm6$ & $1359.6\pm44.11$ & $711.4\pm24.1$ \\
& PGA-MAP-Elites & $1.41{\times}10^5 \pm 6.28{\times}10^3$ & $11.56\pm0.45$ & $289\pm11$ & $\underline{2574.1\pm141.47}$ & $303.5\pm22.8$ \\
& CMA-MAEGA & $5.48{\times}10^4 \pm 2.09{\times}10^3$ & $6.64\pm0.31$ & $166\pm8$ & $1116.7\pm14.66$ & $284.8\pm17.5$ \\
& DNS & $1.20{\times}10^5 \pm 4.96{\times}10^3$ & $7.00\pm0.29$ & $175\pm7$ & $1217.5\pm15.93$ & $461.9\pm19.4$ \\
& QDAC & $\underline{6.06{\times}10^5 \pm 1.92{\times}10^4}$ & $\underline{24.92\pm0.78}$ & $\underline{623\pm20}$ & $1916.3\pm72.21$ & $472.2\pm23.7$ \\
& EDOCS & $1.09{\times}10^5 \pm 4.18{\times}10^3$ & $5.92\pm0.26$ & $148\pm7$ & $1375.8\pm9.64$ & $754.3\pm26.9$ \\
& \textbf{SV-QD-RL} & $\mathbf{8.42{\times}10^5 \pm 2.36{\times}10^4}$ & $\mathbf{31.48\pm0.71}$ & $\mathbf{787\pm18}$ & $\mathbf{4548.2\pm36.5}$ & $\mathbf{1456.3\pm38.4}$ \\
\midrule
\multirow{8}{*}{Ant}
& MAP-Elites & $1.04{\times}10^5 \pm 3.62{\times}10^3$ & $8.56\pm0.33$ & $214\pm8$ & $581.0\pm8.17$ & $283.3\pm14.5$ \\
& NSLC & $1.04{\times}10^5 \pm 3.47{\times}10^3$ & $8.28\pm0.31$ & $207\pm8$ & $570.8\pm5.89$ & $353.0\pm15.9$ \\
& PGA-MAP-Elites & $1.58{\times}10^5 \pm 6.12{\times}10^3$ & $10.28\pm0.44$ & $257\pm11$ & $1511.7\pm101.21$ & $503.2\pm28.4$ \\
& CMA-MAEGA & $2.97{\times}10^5 \pm 8.83{\times}10^3$ & $\underline{19.44\pm0.62}$ & $\underline{486\pm16}$ & $986.6\pm1.48$ & $547.5\pm19.7$ \\
& DNS & $2.58{\times}10^5 \pm 7.94{\times}10^3$ & $17.08\pm0.55$ & $427\pm14$ & $978.3\pm2.18$ & $582.2\pm20.5$ \\
& QDAC & $1.71{\times}10^5 \pm 5.86{\times}10^3$ & $6.56\pm0.27$ & $164\pm7$ & $993.9\pm22.60$ & $541.9\pm18.6$ \\
& EDOCS & $\underline{4.02{\times}10^5 \pm 1.18{\times}10^4}$ & $10.20\pm0.39$ & $255\pm10$ & $\underline{2152.2\pm14.14}$ & $\underline{1527.0\pm31.8}$ \\
& \textbf{SV-QD-RL} & $\mathbf{1.03{\times}10^6 \pm 2.41{\times}10^4}$ & $\mathbf{24.76\pm0.43}$ & $\mathbf{619\pm11}$ & $\mathbf{5171.2\pm32.99}$ & $\mathbf{1571.0\pm27.6}$ \\
\bottomrule
\end{tabular}}
\end{table*}

\begin{figure*}[!t]
\centering
\includegraphics[width=\textwidth]{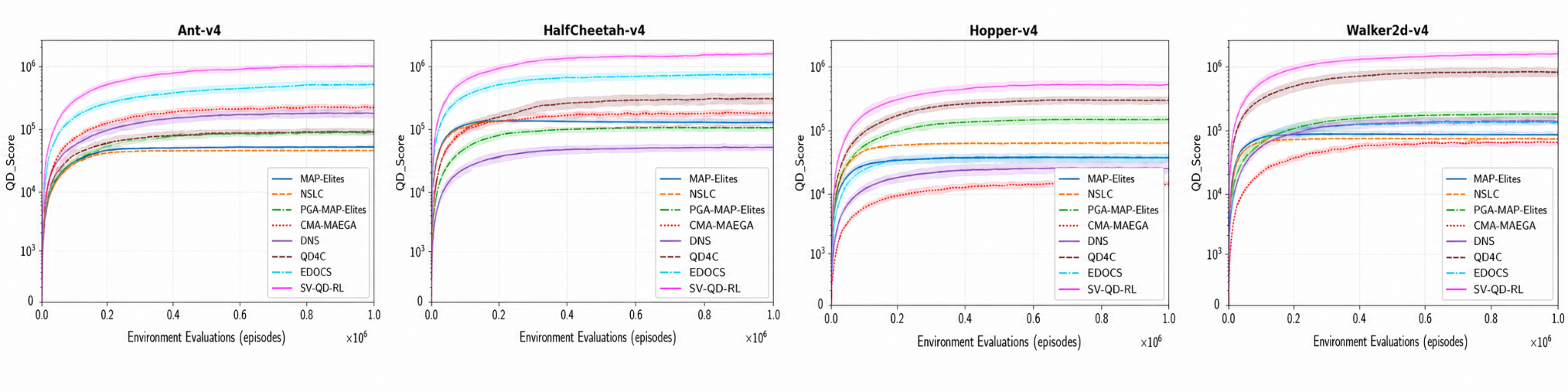}
\caption{QD-score convergence on the four MuJoCo benchmarks. The vertical axis uses log compression with a retained zero baseline; shaded bands denote repeated-run variation.}
\label{fig:qd_convergence}
\end{figure*}

This section reports the main archive-quality and fixed-archive deployment experiments; ablation and mechanism analyses are discussed in Section~\ref{sec:discussion}.

\subsection{Experimental Setup}
\label{sec:exp_setup}

We evaluate SV-QD-RL through comparative experiments across heterogeneous continuous-control locomotion tasks.
The evaluation addresses three core questions.
First, we examine whether SV-QD-RL improves archive quality over classical QD methods, local-competition methods, policy-gradient-assisted QD-RL methods, and recent search-based QD variants.
Second, we analyze whether the learned archive provides stable performance gains across different MuJoCo morphologies under the same interaction budget.
Third, we investigate whether the final archive can serve as a fixed policy library for deployment tasks that require behavior-specific policy selection and backup switching.
The key implementation settings are summarized below, and the deployment-task protocol is described in the supplementary material.

\textbf{Benchmarks and evaluation protocol.}
The evaluation is conducted on four MuJoCo locomotion benchmarks: Hopper-v4, HalfCheetah-v4, Walker2d-v4, and Ant-v4.
These environments cover different body morphologies and contact patterns, including single-legged hopping, planar running, bipedal walking, and multi-legged locomotion.
They therefore provide a suitable benchmark suite for testing whether a method can construct a repertoire of high-performing yet behaviorally distinct policies, rather than optimizing only one high-return controller.

All archive-based methods are evaluated after projection to the same two-dimensional behavior space, $\mathbf{z}(\pi)=[\bar{v}(\pi),\bar{C}_{\mathrm{foot}}(\pi)]$, where $\bar{v}$ is the rollout-average locomotion velocity and $\bar{C}_{\mathrm{foot}}$ is the foot-contact duty factor.
The descriptor range is fixed to $[-1,5]\times[0,1]$ and discretized into a global $50\times50$ grid.
For each method, all final policies are mapped to this shared grid, and each occupied cell retains only the highest-return elite.
We report QD-score, coverage, effective cells, best return, and mean elite return as standard archive-quality indicators; these metrics jointly measure archive quality, explored descriptor area, occupied-cell count, peak performance, and typical elite quality.

\textbf{Compared methods.}
We compare SV-QD-RL with representative baselines from three categories.
The first category includes classical and local-competition QD methods, namely MAP-Elites and NSLC.
The second category includes QD-RL and actor--critic QD methods, including PGA-MAP-Elites, EDOCS, and QDAC.
The third category includes recent search-based or emitter-based QD variants, including CMA-MAEGA and DNS.
These baselines cover grid-based illumination, novelty search with local competition, policy-gradient-assisted variation, actor--critic repertoire learning, evolutionary emitter mechanisms, and diversity-oriented search.

TD3 and SAC are additionally included in the deployment evaluation as single-policy RL baselines.
Since these methods do not construct policy repertoires, they are not treated as archive-quality baselines in the QD-score comparison.
Instead, they are evaluated as one-entry policy libraries to test whether a strong single policy can satisfy diverse speed--contact deployment requirements without archive-level backup candidates.

\textbf{Training and implementation settings.}
All archive-based methods are trained under the same common protocol unless otherwise stated.
The population size is fixed to $100$, and the environment interaction budget is set to $10^6$ steps for each method--environment pair.
Each experiment is repeated with five random seeds, and the reported values are mean $\pm$ standard deviation.
The final metric computation is performed only after projecting the resulting policy set to the shared speed--contact grid.

For SV-QD-RL, each candidate is represented as a structure-conditioned actor--critic branch.
The actor uses a two-layer MLP with $256$ hidden units per layer, while the critic follows the TD3 double-Q architecture with the same $256$--$256$ hidden-layer width.
The main TD3 settings are discount factor $0.99$, target-network coefficient $0.005$, batch size $100$, actor and critic learning rates $10^{-3}$, policy noise $0.2$, noise clip $0.5$, and policy delay $2$.
Each run uses $15$ global iterations, up to $10$ branch-private critics or fine-tuned branches per iteration, $20000$ TD3 refinement steps for each selected branch, a global replay buffer of $10^6$ transitions, and a local branch replay buffer of $10^5$ transitions.
Structural masks are sampled within the target sparsity range $[0.2,0.99]$ and grouped by sparsity-tier boundaries $[0,0.2,0.5,0.7,0.9,1.01]$.
Branches are organized into sparsity tiers, archive admission considers reward, behavioral novelty, structural separation, and value-profile novelty, and NBC-based continuation selects promising archive regions for further refinement.

\textbf{Fixed-archive deployment protocol.}
Deployment evaluation treats the final archive as a frozen policy library.
Each task queries a target region in the speed--contact descriptor space and permits one backup attempt from a fresh simulator state if the first policy fails.
The main metric is deployment success rate, while the candidate-query rule, reset rule, and success-rate definition are given in the supplementary material.

\subsection{QD-RL Archive Performance}
\label{sec:exp_qd}

Table~\ref{tab:main_results} summarizes the final archive state after all methods are projected onto the same speed--contact grid.  Under this common projection, SV-QD-RL gives the strongest overall archive: it combines high QD-score with the largest occupied descriptor area, the largest number of effective cells, and the best elite quality in all four environments.  The table is used here to compare the final repertoires rather than to describe the training process.

The final-table comparison also shows why a single archive metric is insufficient.  Some baselines obtain broad coverage but low mean elite return, while others retain high-return policies in narrower behavior regions.  SV-QD-RL improves both sides because structure-conditioned branches and branch-private critics refine multiple behavior regions instead of concentrating the search around one dominant controller.

Unlike Table~\ref{tab:main_results}, which reports endpoint archive quality, Fig.~\ref{fig:qd_convergence} shows the optimization dynamics and highlights when each method saturates.  MAP-Elites and NSLC grow quickly in the early stage and then plateau, which indicates that descriptor discovery alone is not enough once elite quality becomes the bottleneck.  Actor--critic and emitter-based methods continue improving for longer, while SV-QD-RL keeps a later-phase improvement trend because archive selection, critic-guided branch refinement, and structural proposals remain active throughout training.

\subsection{Deployment Evaluation}
\label{sec:exp_deployment}

\begin{figure}[!t]
\centering
\includegraphics[width=\columnwidth]{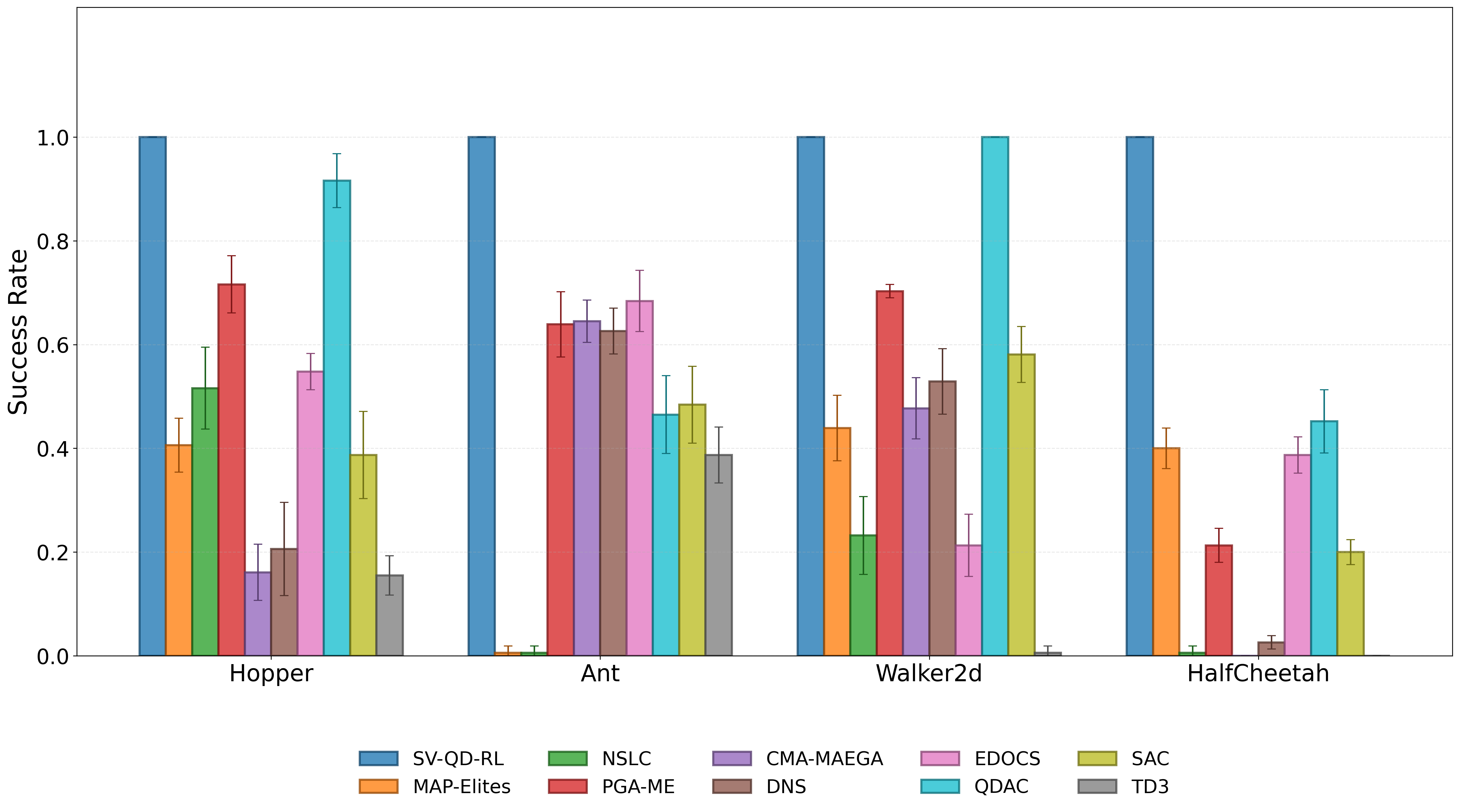}
\caption{Deployment success with one backup switch under the fixed-archive speed--contact protocol.}
\label{fig:deployment_success_rate}
\end{figure}

The second experiment evaluates whether the final archive can be used as a deployment-time policy library after training has stopped.
All methods are frozen during evaluation.
The reported metric is success rate under the one-backup setting: a task succeeds only if the selected policy or one backup policy completes the required distance while satisfying the target speed--contact constraint.
The full deployment-task definition, candidate-query rule, reset rule, and success-rate equation are provided in the supplementary material.

Fig.~\ref{fig:deployment_success_rate} focuses on the downstream use of the archive rather than on archive construction metrics.  SV-QD-RL is the only method that consistently supplies a task-compatible primary policy and a useful backup under the frozen-library protocol.  TD3 and SAC can solve behaviors close to their learned mode but have no archive-level alternatives when the queried descriptor changes.  QD baselines may cover descriptor cells but still fail deployment if the elites in the queried region are weak.  The deployment figure therefore evaluates practical library usability, while Table~\ref{tab:main_results} evaluates final archive quality.

\section{Discussion}
\label{sec:discussion}

\subsection{Ablation Study}
\label{sec:discussion_ablation}

Table~\ref{tab:discussion_ablation_results} isolates the contribution of each module on Hopper-v4.  All variants use the same training budget, archive resolution, descriptor definition, evaluation protocol, and deployment-task sampler as the full method, and only one component is removed at a time.  Specifically, the structure ablation replaces mask search with a fixed dense actor branch, the branch-critic ablation uses a shared critic for all candidate branches, the memory-matching ablation refines selected branches from the global replay pool instead of their matched branch memories, the value-profile ablation removes $d_V$ and $b_{\mathrm{val}}$ from archive admission, and the NBC ablation selects continuation branches by archived return only.  This one-factor design keeps the comparison focused on how each component affects archive construction and downstream deployment.  The table reports the key outcomes needed to compare complete and ablated variants.

\begin{table}[!h]
\centering
\caption{Hopper-v4 ablation study. Coverage is measured by unified-grid occupancy.}
\label{tab:discussion_ablation_results}
\scriptsize
\setlength{\tabcolsep}{1.8pt}
\renewcommand{\arraystretch}{1.02}
\resizebox{\columnwidth}{!}{%
\begin{tabular}{lcccc}
\toprule
\textbf{Variant} & \textbf{QD-Score} & \textbf{Mean Elite} & \textbf{Coverage} & \textbf{Deploy} \\
\midrule
Full SV-QD-RL & $\mathbf{3.62\times10^{5}}$ & \textbf{1718.6} & \textbf{16.24} & $\mathbf{100.0\pm0.0}$ \\
w/o Structure & $1.63\times10^{5}$ & 681.7 & 11.17 & $74.7\pm5.5$ \\
w/o Branch Critic & $1.44\times10^{5}$ & 429.1 & 12.18 & $70.9\pm6.6$ \\
w/o Memory Matching & $2.57\times10^{5}$ & 1277.1 & 10.66 & $80.9\pm5.3$ \\
w/o Value Profile & $1.87\times10^{5}$ & 1011.5 & 8.63 & $80.9\pm5.0$ \\
w/o NBC & $2.04\times10^{5}$ & 667.3 & 14.21 & $74.7\pm5.5$ \\
\bottomrule
\end{tabular}}
\end{table}

The ablations show that the five design choices are complementary.  Removing structure search strongly reduces archive quality and deployment success, which indicates that masks act as behavior-conditioning variables rather than only compression variables.  Without structure search, candidate branches differ mainly by parameter perturbations and critic updates, so many proposals return to similar actor subspaces and the archive loses useful alternatives.  Sharing one critic across branches causes the largest drop in elite quality because different masked actors then receive gradients from the same value surface; this weakens the intended branch-level specialization and makes high-return elites harder to maintain.  Removing memory matching produces a milder but consistent degradation: the branch actor and critic are still private, but their updates are computed from transitions collected by other branches, which blurs the link between structure, replay distribution, and value gradient.  Removing the value profile mainly hurts coverage and deployment because the archive can no longer distinguish branches whose rollout descriptors are close but whose critics encode different local value geometries.  Finally, replacing NBC with top-return selection weakens mean elite quality and deployment reliability, suggesting that continuation should preserve multiple local basins instead of following only the current best-return region.

These results explain the role of each module in the full pipeline.  Mask search first creates alternative actor subspaces; branch-private critics then provide different local improvement directions inside those subspaces; memory matching keeps the critic update tied to the data generated by the same branch; value-profile novelty makes hidden critic differences visible to archive admission; and NBC chooses continuation parents from multiple archive neighborhoods.  The full method performs best because these steps operate sequentially rather than independently: a branch is generated with a structural condition, improved under its own value-learning state, evaluated by behavior and value information, and then either retained or used as a parent for further search.

\subsection{Structure--Value Coupling Analysis}
\label{sec:discussion_structure_value}

\begin{figure*}[!t]
\centering
\subfloat[Ant]{\includegraphics[width=0.48\textwidth]{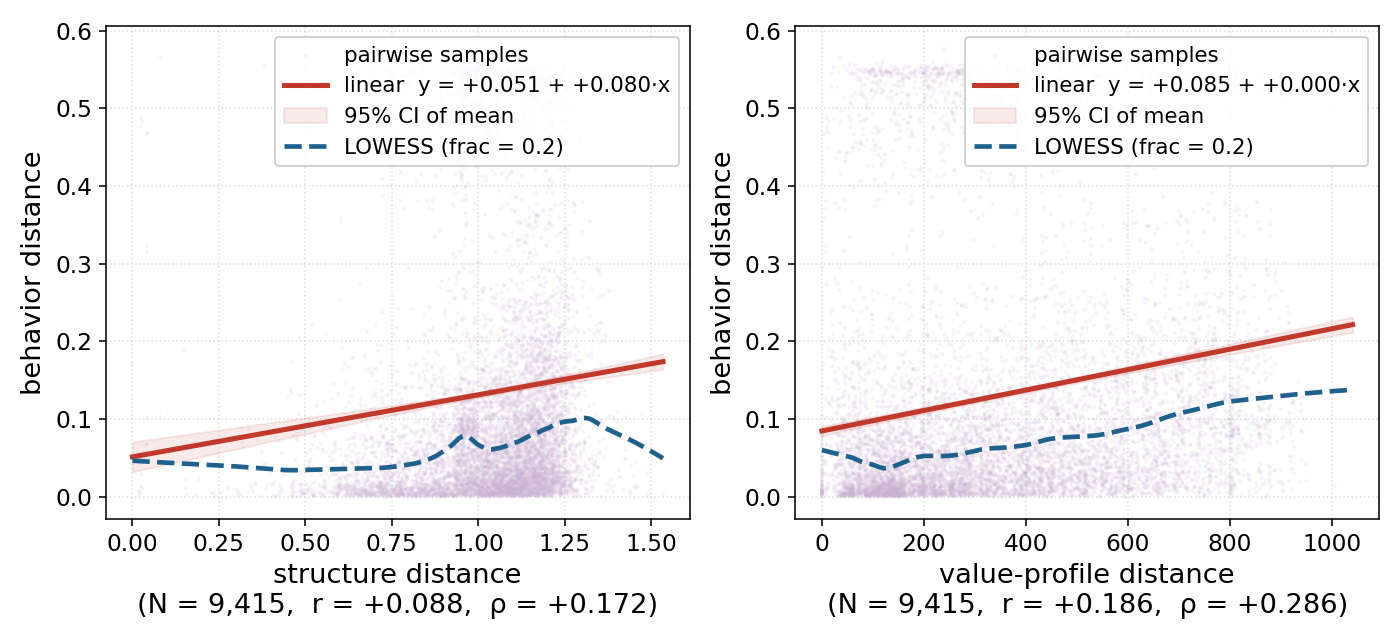}}
\hfill
\subfloat[HalfCheetah]{\includegraphics[width=0.48\textwidth]{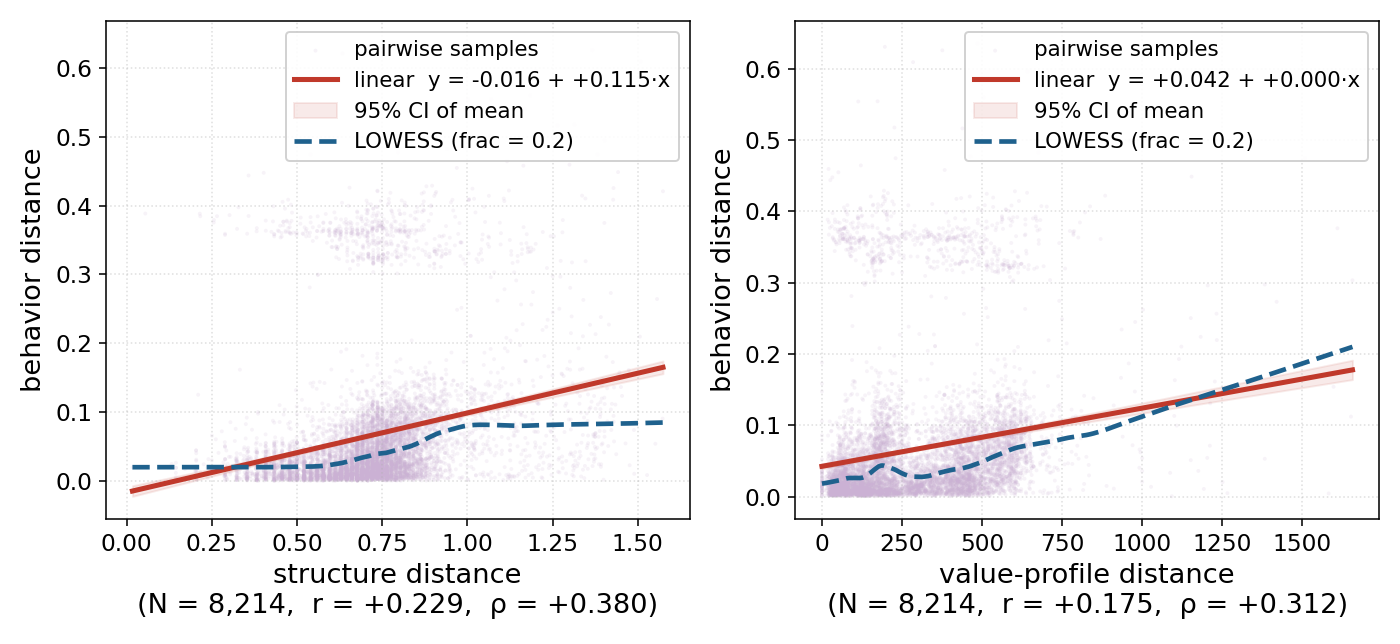}}\\[-0.2em]
\subfloat[Hopper]{\includegraphics[width=0.48\textwidth]{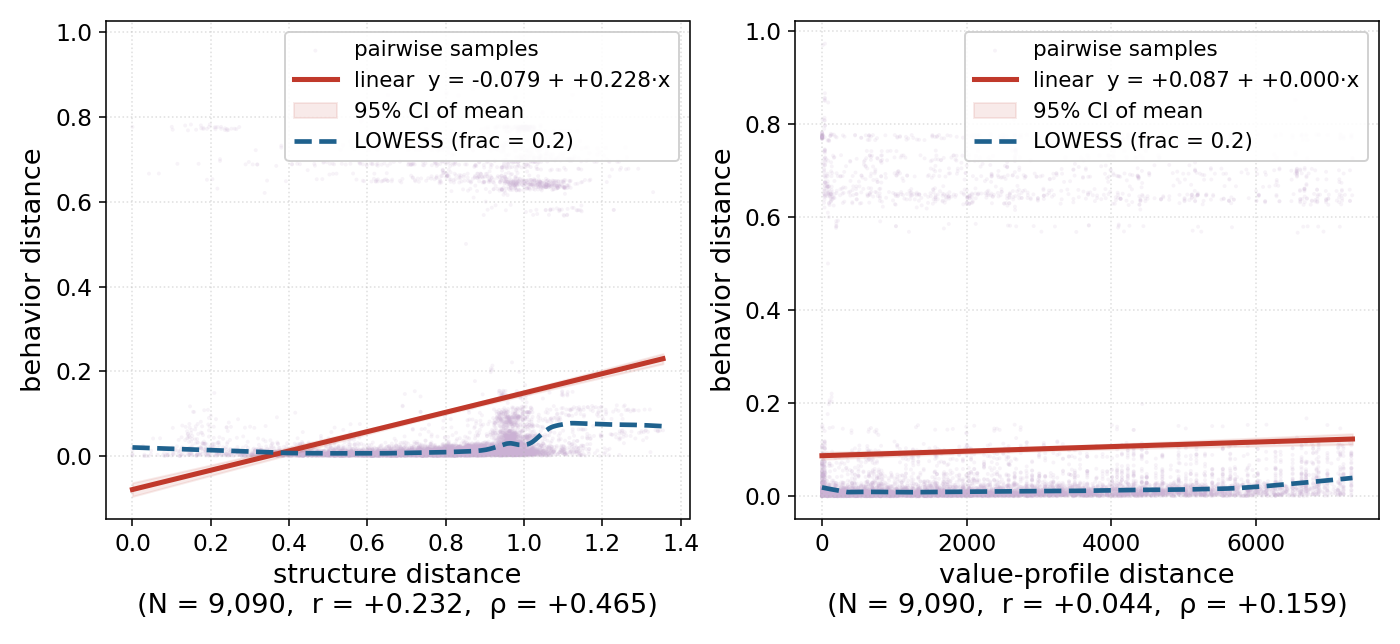}}
\hfill
\subfloat[Walker2d]{\includegraphics[width=0.48\textwidth]{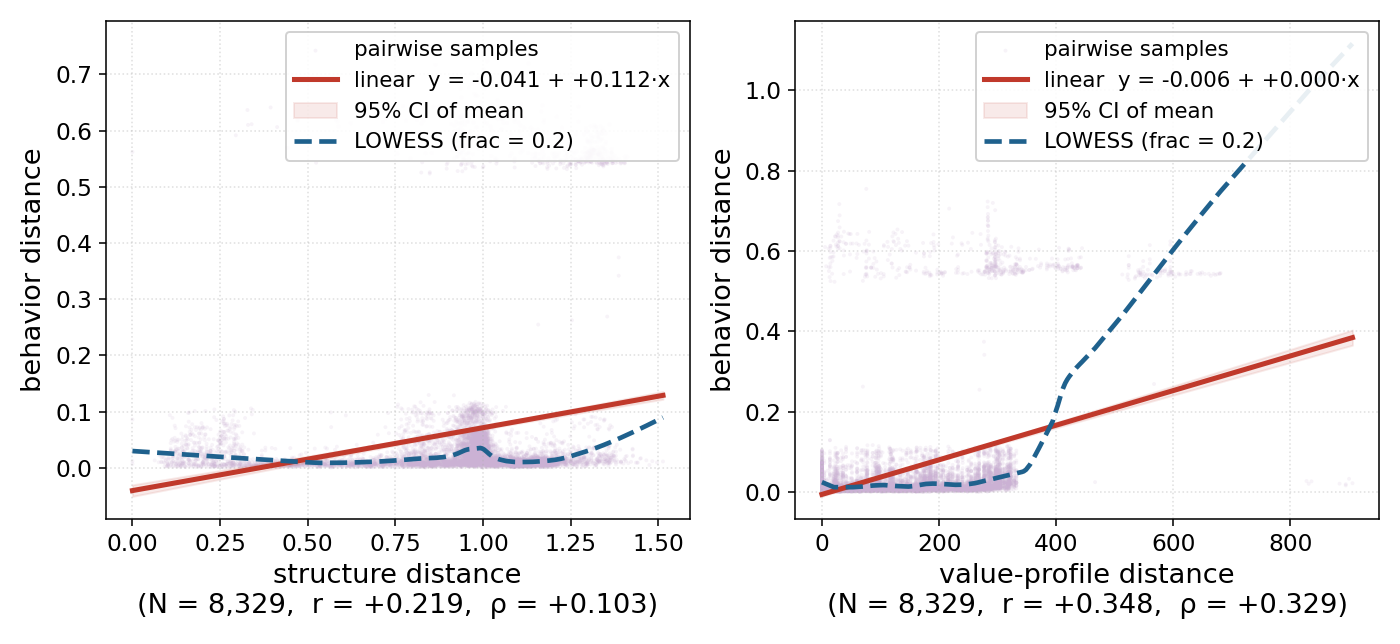}}
\caption{Same-sparsity controlled structure--value coupling. Each panel keeps only branch pairs satisfying $|s_i-s_j|<0.01$ and plots behavior distance against actor-output distance (left) and value-profile distance (right). Linear fits, 95\% mean confidence bands, and LOWESS curves are shown.}
\label{fig:main_same_sparsity_pairwise}
\end{figure*}

The ablation table shows that structure search and branch-private critics are both important, but it does not by itself explain why the final archive contains behaviorally different policies.  We therefore analyze the final archive at the branch-pair level by computing behavior distance $d_B$, actor-output distance $d_S$ on a shared state batch, and critic value-profile distance $d_V$ on a shared state--action batch.  Positive relations between $d_B$ and the other two distances indicate that actor subspaces and critic value geometries are related to behavior separation.

A possible confounding factor is sparsity itself: two policies may behave differently simply because they use substantially different pruning budgets.  To remove this coarse effect, we use a strict same-sparsity slice,
\begin{equation}
\mathcal{P}_{\mathrm{same}}=
\{(i,j):i<j,\; |s_i-s_j|<0.01\},
\label{eq:discussion_same_sparsity_pairs}
\end{equation}
where $s_i$ is the sparsity ratio of branch $i$.  This control keeps only pairs with nearly identical sparsity and therefore tests whether structure and critic differences still explain behavior separation after the pruning level is fixed.

Fig.~\ref{fig:main_same_sparsity_pairwise} provides a branch-pair-level view that is complementary to the ablation results.  Behavior distance remains positively associated with both actor-output distance and value-profile distance after sparsity is fixed, and the LOWESS curves are mostly monotonic in the data-dense regions.  Hopper is more structure dominated, Walker2d and Ant are more value-profile dominated, and HalfCheetah lies between the two cases.  These patterns support the claim that SV-QD-RL does not rely on a single diversity source: masks define actor subspaces, while branch-private critics define behavior-relevant value geometries.

\section{Conclusion}
\label{sec:conclusion}

This paper presented SV-QD-RL, a structure--value coupled framework for constructing deployment-oriented reinforcement learning policy repertoires.  Instead of treating the archive as a passive storage structure, SV-QD-RL uses structural mask search, branch-specific critics, memory-consistent refinement, value-profile novelty, and NBC-based continuation to generate and preserve complementary policies.  Experiments on four MuJoCo locomotion environments show that the method produces high-quality archives and reaches strong deployment success under speed--contact task requirements.  The ablation study and structure--value coupling analysis in Section~\ref{sec:discussion} further support the interpretation that critic-guided structural branching improves repertoire diversity and deployment robustness.

Future work will extend SV-QD-RL to more realistic robotic and edge-control scenarios.  Important directions include state-continuous policy switching, more expressive value-profile measurements using deeper critic layers, and more efficient branch-pruning strategies that preserve archive complementarity while reducing storage and computation.

\bibliographystyle{IEEEtran}
\bibliography{references}

\startsupplementarydocument

\section{Fixed-Archive Deployment Evaluation}
\label{app:deployment_protocol}

This supplementary section provides the deployment-task definition, candidate-query rule, reset rule, and success-rate metric that are summarized only briefly in the main experimental section.

\subsection{Task definition and physical meaning}
\label{app:deployment_task_physical_meaning}

The deployment experiment evaluates whether a trained archive can be used as a fixed policy library after learning has stopped.
No gradient update, archive insertion, or online adaptation is allowed during deployment.
A deployment task is written as
\begin{equation}
\tau=(\mathbf{b}^{\star},D,\epsilon,K),
\label{eq:app_deploy_task}
\end{equation}
where $\mathbf{b}^{\star}$ is the target behavior descriptor, $D$ is the required travel distance, $\epsilon$ is the descriptor tolerance, and $K$ is the maximum number of allowed policy switches.
The main paper reports the one-switch setting, i.e., $K=1$.

The two descriptor dimensions have direct physical interpretations.
The velocity dimension $\bar{v}$ measures task progress: a high value corresponds to a policy that covers distance quickly, while a low value corresponds to slow or cautious locomotion.
The foot-contact duty factor $\bar{C}_{\mathrm{foot}}$ measures the fraction of time in which the robot maintains ground contact through its task-specific foot or foot-like contact sensors.
A high contact duty factor usually indicates a more grounded or stable gait, while a low contact duty factor indicates a more dynamic or contact-light gait.
Thus, the speed--contact descriptor does not only serve as an abstract QD coordinate; it describes a physically meaningful tradeoff between progress rate and contact pattern.

The four corner regions of the descriptor rectangle correspond to four deployment families.
A slow/light-contact target asks for conservative motion with limited ground-contact time, which can represent contact-sensitive motion or recovery-like behavior.
A slow/stable-contact target asks for cautious locomotion with frequent ground support, which is relevant when stability is more important than speed.
A fast/dynamic-contact target asks for rapid movement with relatively short contact phases, which resembles running or bounding behavior.
A fast/stable-contact target asks for high progress rate while maintaining frequent support, which is useful when the robot must move quickly without sacrificing contact robustness.
By sampling target descriptors from these regions, the deployment test evaluates whether the archive contains policies that are not only high-return but also selectable for different physical behavior requirements.

\subsection{Candidate selection and success criterion}
\label{app:deployment_candidate_selection}

For a fixed archive $\mathcal{A}$ and a target descriptor $\mathbf{b}^{\star}$, candidate policies are obtained by a descriptor-radius query,
\begin{equation}
\mathcal{S}(\mathbf{b}^{\star})=
\left\{\pi_i\in\mathcal{A}:\left\|\tilde{\mathbf{b}}_i-\tilde{\mathbf{b}}^{\star}\right\|_2\le \epsilon\right\},
\label{eq:app_candidate_set}
\end{equation}
where $\tilde{\mathbf{b}}$ denotes the descriptor normalized by the global descriptor range.
For grid-based archives, this query is implemented by searching the target cell and neighboring cells whose normalized descriptor distance is within $\epsilon$.
For local-competition or unstructured archives, the same rule is implemented as a circular query in descriptor space.
The selected candidates are sorted by archived return,
\begin{equation}
\pi_{(1)},\pi_{(2)},\ldots=
\operatorname{sort}_{\pi_i\in\mathcal{S}(\mathbf{b}^{\star})}
\left(J_i\right),
\label{eq:app_candidate_ranking}
\end{equation}
with higher-return candidates tested first.
If $\mathcal{S}(\mathbf{b}^{\star})$ is empty, the attempt fails.

A policy attempt succeeds only if one policy independently completes the required distance $D$ from a fresh initial state and remains within the descriptor tolerance.
If the first selected policy fails, the evaluator may test one unused backup policy from the same target region in the $K=1$ setting.
Each backup attempt starts from a fresh simulator state, and the travel distance achieved by failed attempts is discarded.
This reset rule is important physically: after a fall, unstable contact, or task violation, the next policy should not receive credit for distance already covered by a failed controller.
The protocol therefore evaluates fallback among complete policies rather than trajectory stitching.

For task $n$, the success indicator is
\begin{equation}
S_n=\mathbb{I}\left[
\exists j\in\{1,\ldots,K+1\}:\
\pi_{n,j}\ \text{completes}\ D_n\ \text{and satisfies}\ \epsilon_n
\right].
\label{eq:app_deploy_success_indicator}
\end{equation}
The success rate is
\begin{equation}
\mathrm{SR}(K)=\frac{1}{N_{\mathrm{task}}}\sum_{n=1}^{N_{\mathrm{task}}}S_n.
\label{eq:app_success_rate}
\end{equation}
The reported evaluation samples $N_{\mathrm{task}}=31$ tasks per environment and repeats task sampling five times.
TD3 and SAC use the same task set but contain only one candidate policy, so they cannot benefit from the backup switch.

\end{document}